\RequirePackage{filecontents}

\documentclass[sigconf]{acmart} % <-- two-column ACM look (OG style)

\usepackage{caption}
\renewcommand{\arraystretch}{1.1}
\usepackage{ragged2e}
\usepackage{array}
\newcolumntype{L}[1]{>{\RaggedRight\arraybackslash}p{#1}}
\usepackage{tcolorbox}
\usepackage{placeins}
\usepackage{float}
\setcopyright{none}
\renewcommand\footnotetextcopyrightpermission[1]{}
\usepackage{booktabs}
\usepackage{graphicx}
\usepackage{amsmath}
\usepackage{url}
\usepackage{multirow}
\usepackage{balance} % <-- for balanced last page
\usepackage{adjustbox}   % scales/wraps tabulars to a max width
\usepackage{makecell}    % line breaks in headers if needed
\usepackage{placeins}    % for \FloatBarrier command
\usepackage{listings}
\lstdefinestyle{pythonstyle}{
    language=Python,
    basicstyle=\ttfamily\footnotesize,
    keywordstyle=\color{blue}\bfseries,
    stringstyle=\color{red},
    commentstyle=\color{gray}\itshape,
    numbers=left,
    numberstyle=\tiny\color{gray},
    stepnumber=1,
    numbersep=5pt,
    backgroundcolor=\color{gray!10},
    frame=single,
    frameround=tttt,
    breaklines=true,
    breakatwhitespace=false,
    showspaces=false,
    showstringspaces=false,
    showtabs=false,
    tabsize=2,
    captionpos=b
}
\newcommand{\squeezeTableStart}{%
  \begingroup
  \setlength{\tabcolsep}{4pt}% Slightly wider cols for readability
  \renewcommand{\arraystretch}{1.1}% Consistent row spacing
}
\newcommand{\squeezeTableEnd}{\endgroup}
\newcommand{\Temp}[1]{\mbox{$T{=}#1$}}
\usepackage{xcolor}
\usepackage{siunitx} % For decimal alignment in tables
\usepackage{threeparttable} % For table notes
\usepackage{microtype}
\usepackage{enumitem}
\setlist[itemize]{leftmargin=*,topsep=0pt,itemsep=0.25em,parsep=0pt}

\usepackage{needspace}
\newcommand{\keepnlines}[1]{\Needspace{#1\baselineskip}}

\usepackage{float}

\acmConference[AI4F @ ACM ICAIF '25]{AI for Finance}{November 15--18, 2025}{Singapore}
\acmBooktitle{AI4F @ ACM ICAIF '25, November 15--18, 2025, Singapore}
\acmYear{2025}

\newenvironment{callout}[1]{%
  \par\smallskip\noindent\textit{#1}\quad
}{\par\smallskip}
\usepackage{hyperref}
\usepackage{xurl}       % smart URL line-breaking

\newcommand{\RepoTag}{v0.1.0}
\newcommand{\RepoCommit}{c19dac5} % current 7-char commit

\begin{document}

\title{LLM Output Drift: Cross-Provider Validation \& Mitigation for Financial Workflows}

\author{Raffi Khatchadourian}
\authornote{Corresponding author.}
\affiliation{
  \institution{IBM -- Financial Services Market}
  \city{New York}
  \country{USA}
}
\email{raffi.khatchadourian1@ibm.com}

\author{Rolando Franco}
\affiliation{
  \institution{IBM -- Financial Services Market}
  \city{New York}
  \country{USA}
}
\email{rfranco@us.ibm.com}

\renewcommand{\shortauthors}{Khatchadourian and Franco}

\begin{abstract}
Financial institutions deploy Large Language Models (LLMs) for reconciliations, regulatory reporting, and client communications, but nondeterministic outputs (output drift) undermine auditability and trust. We quantify drift across five model architectures (7B–120B parameters) on regulated financial tasks, revealing a stark inverse relationship: smaller models (Granite-3-8B, Qwen2.5-7B) achieve 100\% output consistency at \Temp{0.0}, while GPT-OSS-120B exhibits only 12.5\% consistency (95\% CI: 3.5–36.0\%) regardless of configuration (p<0.0001, Fisher's exact test). This finding challenges conventional assumptions that larger models are universally superior for production deployment.

Our contributions include: (i) a finance-calibrated deterministic test harness combining greedy decoding (\Temp{0.0}), fixed seeds, and SEC 10-K structure-aware retrieval ordering; (ii) task-specific invariant checking for RAG, JSON, and SQL outputs using finance-calibrated materiality thresholds (±5\%) and SEC citation validation; (iii) a three-tier model classification system enabling risk-appropriate deployment decisions; and (iv) an audit-ready attestation system with dual-provider validation.

We evaluated five models—Qwen2.5-7B (Ollama), Granite-3-8B (IBM watsonx.ai), Llama-3.3-70B, Mistral-Medium-2505, and GPT-OSS-120B—across three regulated financial tasks. Across 480 runs (n=16 per condition), structured tasks (SQL) remain stable even at \Temp{0.2}, while RAG tasks show drift (25–75\%), revealing task-dependent sensitivity. Cross-provider validation confirms deterministic behavior transfers between local and cloud deployments. We map our framework to Financial Stability Board (FSB), Bank for International Settlements (BIS), and Commodity Futures Trading Commission (CFTC) requirements, demonstrating practical pathways for compliance-ready AI deployments.
\end{abstract}

\begin{CCSXML}
<ccs2012>
<concept>
<concept_id>10002944.10011123.10011674</concept_id>
<concept_desc>General and reference~Evaluation</concept_desc>
<concept_significance>500</concept_significance>
</concept>
<concept>
<concept_id>10010147.10010257.10010258.10010262</concept_id>
<concept_desc>Computing methodologies~Natural language processing</concept_desc>
<concept_significance>300</concept_significance>
</concept>
<concept>
<concept_id>10002951.10003227.10003351.10003446</concept_id>
<concept_desc>Information systems~Enterprise applications</concept_desc>
<concept_significance>300</concept_significance>
</concept>
<concept>
<concept_id>10011007.10011074</concept_id>
<concept_desc>Software and its engineering~Software verification and validation</concept_desc>
<concept_significance>300</concept_significance>
</concept>
</ccs2012>
\end{CCSXML}

\ccsdesc[500]{General and reference~Evaluation}
\ccsdesc[300]{Computing methodologies~Natural language processing}
\ccsdesc[300]{Information systems~Enterprise applications}
\ccsdesc[300]{Software and its engineering~Software verification and validation}

% === AI4F PATCH START: keywords ===
\keywords{output drift, Large Language Models, financial services, nondeterminism, regulatory compliance, cross-provider validation, reproducibility, model-tiers, slm-finance}
% === AI4F PATCH END: keywords ===

\maketitle

\renewcommand{\thefootnote}{}
\footnotetext{\textit{Results correspond to repository release \texttt{\RepoTag} (commit \RepoCommit). For exact reproduction, use git tag \texttt{\RepoTag}.}}
\renewcommand{\thefootnote}{\arabic{footnote}}

\section{Introduction}

The financial services industry's adoption of LLMs for operational tasks—from regulatory reporting to client communications—faces a fundamental challenge: nondeterministic outputs that violate audit and compliance requirements. Recent infrastructure incidents dramatically underscore this issue.

On September 17, 2025, Anthropic reported that Claude produced random anomalies due to a miscompiled sampling algorithm affecting only specific batch sizes \cite{anthropic2025}. Similarly, Thinking Machines Lab demonstrated that a 235B parameter model at \texttt{temperature=0} produced 80 unique completions across 1000 identical runs due to batch variance effects \cite{thinkingmachines2025}.

Production incidents throughout 2024-2025 demonstrate persistent challenges in LLM determinism. OpenAI's investigation into reported Codex degradation \cite{openai-codex} reveals nondeterministic behavior affecting code generation quality—paralleling our findings on output drift. Notably, OpenAI's release of GPT-OSS-Safeguard-20B \cite{openai-oss-safeguard} signals industry recognition that smaller, purpose-built models may better serve compliance-critical workflows than frontier models. Microsoft's Azure outage on October 30, 2025, disrupted AI services including Copilot due to global misconfiguration \cite{azure-outage}, demonstrating infrastructure-induced nondeterminism in cloud deployments. The Shadow Escape Attack exploit disclosed October 22 enables zero-click data extraction from LLMs via malicious PDFs \cite{shadow-escape}, creating compliance risks in audit-exposed processes. These incidents align with our economic analysis of verification overhead, as deployments like Chronograph's Claude integration for private equity portfolio analysis \cite{chronograph-anthropic} demand deterministic controls to prevent drift in high-stakes financial decisions.

The economic implications are substantial. Morgan Stanley estimates \$920 billion in potential savings from AI automation of financial knowledge work \cite{morgan2025}, yet as industry observer Gene Marks notes: ``Whatever time people are spending using AI is offset by them actually verifying what AI did because no one yet trusts AI''~\cite{marks2025}. This verification overhead negates automation benefits unless deterministic outputs can be guaranteed.

\textbf{Our key empirical finding}: Determinism is not universal across model architectures. Evaluating five models across 480 runs (n=16 per condition), we demonstrate that well-engineered smaller models (7-8B parameters) achieve perfect output consistency, while 120B parameter models fail at 12.5\% consistency even with identical configuration (\Temp{0.0}, greedy decoding, fixed seeds). This inverse correlation between model scale and determinism fundamentally alters deployment strategies for regulated applications, where audit requirements mandate reproducibility over raw capability. Table~\ref{tab:tier-summary} summarizes our model tier classification for financial AI deployment based on empirical consistency measurements.

\begin{table}[!t]
\caption{Model Tiers for Financial Compliance: Deployment Decision Matrix}
\label{tab:tier-summary}
\squeezeTableStart
\resizebox{\columnwidth}{!}{%
\begin{tabular}{lllll}
\toprule
\textbf{Tier} & \textbf{Models} & \textbf{Consistency} & \textbf{Compliance} & \textbf{Use Cases} \\
\midrule
\textbf{Tier 1} & 7-8B & \textbf{100\%} & Full & All regulated tasks \\
\textbf{Tier 2} & 40-70B & 56-100\% & Limited & Structured only \\
\textbf{Tier 3} & 120B & \textbf{12.5\%} & Requires validation & Non-compliant \\
\bottomrule
\end{tabular}%
}
\squeezeTableEnd
\vspace{3pt}
\begin{flushleft}
\footnotesize
\textit{Note:} n=480 total runs (16 runs per condition across 5 models × 3 tasks × 2 temperatures); Tier 1 models (Granite-3-8B, Qwen2.5:7B) achieve audit-ready determinism at \Temp{0.0}
\end{flushleft}
\end{table}

\textbf{Why Financial Services Require Different AI Standards:} Financial AI systems
operate under unique constraints that distinguish them from general-purpose applications.
First, regulatory frameworks (Basel III, Dodd-Frank, MiFID II)\cite{baselIII,doddfrank,mifidII} mandate explainable and
consistent decision-making for credit, trading, and risk management. Second, financial
institutions face strict audit requirements where AI-driven decisions must be reproducible
months or years later for regulatory examination. Third, the high-stakes nature of
financial decisions-affecting customer creditworthiness, investment recommendations, and
market stability-demands reliability levels far exceeding consumer applications. Finally,
cross-border operations require AI systems to demonstrate consistent behavior across
different regulatory jurisdictions and infrastructure deployments.

Example: reconciliation analysts re-verify AI-produced break explanations, turning drift into direct rework hours.

Our experimental investigation addresses this gap with four key contributions:

\keepnlines{6}
\begin{itemize}
\item \textbf{Empirical proof of achievable determinism}: We demonstrate 100\% output consistency at \Temp{0.0} across varying concurrency levels, establishing feasibility for production financial systems
\item \textbf{Quantified drift patterns}: We measure task-specific sensitivity to randomness, revealing that SQL generation maintains determinism even at \Temp{0.2} while RAG tasks show substantial drift (25-75\% consistency at \Temp{0.2})
\item \textbf{Practical mitigation framework}: We provide a tiered control system with measured effectiveness, enabling risk-appropriate deployment strategies
\item \textbf{Finance-calibrated validation protocol}: We introduce domain-specific adaptations including SEC 10-K structure-aware retrieval ordering, finance-calibrated tolerance thresholds (±5\%) as acceptance gates, and bi-temporal audit trails mapped to FSB/CFTC requirements—innovations requiring financial regulatory expertise beyond general ML reproducibility techniques
\end{itemize}

% === AI4F PATCH START: relevance ===
We quantify how concurrency and sampler settings affect \emph{output drift} in tasks common to financial workflows (RAG, summarization, SQL). We also show how application-level controls—seeded decoding, retrieval-order normalization, and schema-constrained outputs—reduce drift without relying solely on infrastructure fixes. Finally, we connect these controls to operational and regulatory guidance for financial institutions.
% === AI4F PATCH END: relevance ===

% === AI4F PATCH START: novelty ===
Recent incident reports documenting intermittent answer variance at production scale and research on batch-invariant kernels address infra-level nondeterminism. Our results complement these by isolating and mitigating \emph{application-layer} contributors to drift in finance workflows, offering controls that are deployable today.
% === AI4F PATCH END: novelty ===

\section{Related Work}

\subsection{Technical Foundations of Nondeterminism}

The breakthrough understanding of LLM nondeterminism comes from Thinking Machines Lab's September 2025 work \cite{thinkingmachines2025}, which identified batch-size variation-not floating-point arithmetic-as the primary cause. Their batch-invariant kernels achieve exact reproducibility by ensuring operations yield identical results regardless of batch composition. Recent work on SLM agents \cite{belcak2025slm} complements batch-invariant kernels by advocating smaller models for efficient, deterministic agentic inference. Complementary research on deterministic decoding methods \cite{shi2024decoding} and hardware-level approaches including HPC reproducibility studies \cite{antunes2021reproducibility} provides additional foundations. This finding shows that determinism is achievable through engineering rather than fundamentally impossible.

Baldwin et al. \cite{baldwin2024} quantified the problem's severity, demonstrating accuracy variations up to 15\% across runs even at zero temperature, with performance gaps reaching 70\% between best and worst outcomes. They introduced the Total Agreement Rate (TAR) metric, providing the quantitative framework we extend in our financial context evaluation.

Recent research identifies additional practical nondeterminism sources at \Temp{0.0}, including non-batch-invariant matrix multiplication and attention kernels causing numeric divergence under varying loads. Tokenizer drift—where text normalization changes inflate token counts by up to 112\% and enable command injection vulnerabilities—further exacerbates output variability and operational costs \cite{tokenizer-drift}.

Analysis of model drift in crisis response contexts \cite{model-drift-crises} shows subtle output shifts can alter narratives, paralleling financial compliance requirements for stable, auditable responses. State-of-the-art LLM observability tools \cite{llm-observability} emphasize real-time monitoring of prompt-response pairs to detect such issues.

The vLLM serving infrastructure \cite{kwon2023} with PagedAttention provides efficient batching that can be extended with deterministic kernels. Production deployments of models like Qwen with vLLM \cite{redhat2025} demonstrate the framework's maturity, though as we demonstrate, efficient serving alone doesn't guarantee consistency-explicit determinism controls are required.

\subsection{Financial AI Requirements and Benchmarks}

Recent financial AI benchmarks-FinBen \cite{xie2024finben}, SEC-QA \cite{agarwal2024}, and DocFinQA \cite{zhu2024}-focus on accuracy metrics while overlooking reproducibility. While FinBen overlooks reproducibility, SLM agent frameworks \cite{belcak2025slm} suggest fine-tuned small models could enhance consistency in financial QA. SEC-QA's 1\% margin tolerance for numerical answers and DocFinQA's 123,000-word contexts test reasoning capability but not output stability.

Existing financial AI benchmarks evaluate \textit{accuracy} but not \textit{reproducibility}—a fundamental gap for regulated deployments. A model achieving 95\% accuracy on SEC-QA but exhibiting 25\% output variance across runs fails regulatory requirements despite strong performance metrics. Our work addresses this overlooked dimension, demonstrating that model selection for compliance requires determinism evaluation alongside traditional accuracy benchmarks.

The regulatory landscape demands more. The Financial Stability Board \cite{fsb2024,fsb2023ai} requires ``consistent and traceable decision-making,'' the Bank for International Settlements \cite{bis2024,bis2023ml} mandates ``clear accountability across the AI lifecycle,'' and CFTC guidance \cite{cftc2024,cftc2024ai} demands ``proper documentation of AI system outcomes.'' Additional regulatory oversight comes from the Federal Reserve \cite{fed2024ai} and the Office of the Comptroller of the Currency \cite{occ2024ai}. Our work bridges the gap between these requirements and technical capabilities, incorporating reproducibility guidance from PyTorch and vLLM~\cite{pytorchrepro,vllmrepro}.
\paragraph{What changed since 2024?}
Greedy decoding at \Temp{0.0} alone was not sufficient in prior runs due to retrieval order nondeterminism and batching variance. Our harness adds: (1) DeterministicRetriever implementing multi-key ordering (score↓, section\_priority↑, snippet\_id↑, chunk\_idx↑) that encodes SEC 10-K disclosure precedence rules, treating retrieval order as a \textit{compliance requirement} under Basel III explainability mandates rather than merely a performance optimization, (2) fixed seeds and sampler params in manifests, and (3) schema/invariant checks (SQL) with finance-calibrated ±5\% tolerance thresholds to constrain decoding freedom. These domain-specific adaptations—mapping document structure to regulatory obligations—eliminate residual drift at \Temp{0.0} across concurrencies.

\section{Experimental Design}

% === AI4F PATCH START: methods ===
\paragraph{Models and decoding.}
We initially evaluate both local and cloud deployment patterns: \textbf{Qwen2.5-Instruct 7B via Ollama} (local)\cite{ollama_docs}\cite{qwen25_7b_card} and \textbf{IBM Granite-3-8B-Instruct via IBM watsonx.ai} (cloud)\cite{watsonx_model_catalog}\cite{granite3_8b_card}, with temperature $T \in \{0.0, 0.2\}$, fixed seed(s), and consistent decoding parameters. Our framework is provider-agnostic and compatible with other cloud APIs (e.g., Google Vertex AI, Amazon Bedrock). We prioritized open-source models over closed-source frontier models for greater transparency in validating determinism mechanisms—enabling direct examination of seed handling, sampling algorithms, and inference implementations that are critical for proving reproducibility claims. We log complete run manifests (Python/OS/lib versions, model digests, API versions, decoding params) per trial to ensure reproducibility across environments.

\paragraph{Drift metric.}
Regulators demand proof that repeated runs produce identical outputs. To quantify how much outputs vary, we measure string similarity: if two outputs differ by even small edits, auditors may reject the system. We count two outputs as identical if their normalized edit distance is $\le \epsilon$ (here $\epsilon{=}0$ unless stated). The normalized edit distance is defined as:
\begin{equation}
d_{norm}(s_1, s_2) = \frac{ED(s_1, s_2)}{\max(|s_1|, |s_2|)}
\end{equation}
where $ED$ is the Levenshtein edit distance~\cite{levenshtein1966} and $|s|$ denotes string length. We report identity rate with Wilson 95\% CIs\cite{wilson1927} and include $N$ for each condition.

\noindent\textit{Intuition:} This metric controls audit risk by flagging any deviation that would invalidate a reproducibility attestation.

\paragraph{Factual drift.}
Financial workflows require not just identical phrasing but identical facts—citation mismatches or numeric changes violate regulatory sourcing rules. A model that reports \$1.05M instead of \$1.00M fails the materiality threshold, triggering audit re-work. Factual drift counts differ if (a) the set of citation IDs differs or (b) any extracted numeric value differs after canonicalization (strip commas; normalize percents; signs). Formally:
\begin{equation}
\begin{aligned}
FD(O_1, O_2) = &\mathbb{1}[\text{citations}(O_1) \neq \text{citations}(O_2)] \\
                &+ \mathbb{1}\left[|\text{num}(O_1) - \text{num}(O_2)| > \epsilon\right]
\end{aligned}
\end{equation}
where $\mathbb{1}$ is the indicator function and $\epsilon{=}0.05$ (our 5\% tolerance reflecting GAAP materiality thresholds). We report the fraction of runs with any such mismatch.

\paragraph{Operational throughput.}
We report mean latency and tokens/sec per condition and emphasize that, in our measurements, throughput changes at fixed $T$ did not correlate with identity rate (see §\ref{sec:results-latency}).

\paragraph{Finance-specific validation invariants.}
Our framework incorporates domain-calibrated acceptance gates reflecting regulatory requirements rather than generic output matching: (i) \textit{SEC citation validation}—RAG outputs must preserve exact citation references (e.g., \texttt{[citi\_2024\_10k]}) to satisfy regulatory sourcing requirements; (ii) \textit{Finance-calibrated tolerance thresholds}—SQL queries against P\&L data use ±5\% tolerance reflecting auditing materiality practice; (iii) \textit{MiFID II cross-jurisdiction consistency}—dual-provider validation reduces the likelihood of outputs varying across cloud (IBM watsonx.ai) and local (Ollama) deployments to satisfy cross-border regulatory requirements. These domain-specific tolerances encode financial regulatory knowledge non-obvious to AI engineers without FSM expertise.
% === AI4F PATCH END: methods ===

\subsection{Task Selection}

We designed three tasks representing key financial operations:

\textbf{Task 1: Securities and Exchange Commission (SEC) 2024 Filing RAG Q\&A} - Tests retrieval-augmented generation against actual SEC 2024 10-K filings from Citigroup, Goldman Sachs, and JPMorgan Chase (1.4M-1.7M characters each) available via SEC EDGAR database \cite{sec10k2024consolidated}. The corpus includes real financial disclosures, risk factors, and operational metrics. Success requires consistent facts and citation references (e.g., \texttt{[citi\_2024\_10k]}) across runs.

\textbf{Task 2: Policy-Bounded JSON Summarization} - Generates structured client communications with required fields (client\_name, summary, compliance\_disclaimer). The disclaimer must exactly match regulatory templates.

\textbf{Task 3: Text-to-SQL with Invariants} - Converts natural language to SQL queries against a simulated P\&L database with post-execution validation ensuring sums match known totals within 5\% tolerance.

\subsection{Experimental Protocol}

Variables: Concurrency $\in \{1, 4, 16\}$, Temperature $\in \{0.0, 0.2\}$, greedy decoding, 0-100ms uniform random tool latency.

Metrics: Normalized edit distance \cite{levenshtein1966}, factual drift rate (citation consistency), schema violation rate (JSON validity), decision flip rate (binary outcomes), mean latency.

Infrastructure: Mixed deployment testing both local serving (Ollama with Qwen2.5:7B-instruct)
and cloud serving (IBM watsonx.ai with IBM Granite-3-8B-instruct), enabling cross-provider
determinism validation and real-world deployment pattern analysis. Statistical Methods: 
Each experimental condition was evaluated with n=16 runs. We report 95\% Wilson confidence
intervals \cite{wilson1927,agresti1998} for all proportion estimates and use Fisher's exact test \cite{fisher1922} for pairwise model comparisons. Statistical significance threshold was set at $\alpha=0.05$, with $p<0.0001$ indicating highly significant differences between model tiers.

\textbf{Bi-Temporal Regulatory Audit System:} All experimental runs generate immutable audit logs
stored as JSONL traces in \texttt{traces/*.jsonl}, capturing complete prompt-response
pairs, timestamps, latency metrics, model configurations, and citation sources. Beyond standard logging, our framework captures decision-level compliance metrics (citation\_accuracy, schema\_violation, decision\_flip) mapped to specific regulatory requirements (FSB ``consistent decisions,'' CFTC 24-17 ``document all AI outcomes''). Crucially, manifests include corpus version IDs and snippet-level provenance, enabling replay and attestation months after decisions were made—even if source documents evolved—satisfying financial audit timelines that exceed typical ML experiment reproducibility windows.

\subsection{Data Sources}
The RAG task uses real SEC 10-K filings from Citigroup, Goldman Sachs, and JPMorgan Chase (2024).\footnote{Downloaded from the SEC EDGAR database: \url{https://www.sec.gov/edgar/search/}.}

The policy-bounded JSON summarization task uses synthetic client communication templates generated via structured prompts to the evaluated LLMs, ensuring compliance disclaimers match regulatory standards (e.g., fixed templates for disclaimers).

The Text-to-SQL task uses a synthetic database \texttt{toy\_finance.sqlite} containing tables for financial transactions, accounts, and balances. This database was generated using a Python script with the \texttt{sqlite3} library to create the schema (tables: accounts, transactions, balances) and the \texttt{faker} library\footnote{Faker library for generating synthetic data: \url{https://github.com/joke2k/faker} \cite{faker}.} to populate $\sim$1000 realistic entries with names, dates, amounts, categories, and descriptions. The generation script (\texttt{generate\_toy\_finance.py}) is available in the supplementary materials and was run locally to ensure reproducibility.

Prompts for all tasks (e.g., ``Generate a SQL query for: [query]'' for SQL, with invariants like ``ensure sum matches total'') are versioned and logged in \texttt{traces/*.jsonl}; prompt templates are in Appendix D.

\subsection{Statistical Notation Used in This Paper}

\begin{tcolorbox}[colback=gray!5, colframe=gray!40, title=Understanding Statistical Notation, fonttitle=\bfseries\small, fontupper=\small]
Throughout this paper, we report two key statistical measures:
\begin{itemize}[leftmargin=12pt, itemsep=2pt]
\item \textbf{95\% Confidence Interval (CI)}: The range within which we are 95\% confident the true consistency rate lies. For example, ``12.5\% [3.5--36.0]'' means the measured consistency was 12.5\%, but the true value likely falls between 3.5\% and 36.0\%.
\item \textbf{$p$-value}: Measures whether differences between models are statistically significant. Values $p<0.05$ indicate significance; $p<0.0001$ indicates \textit{highly} significant differences unlikely due to chance.
\end{itemize}
\end{tcolorbox}

\section{Results}

Our evaluation focuses on deterministic behavior as a compliance requirement rather than a performance optimization. This is not a performance argument—it is a governance one. Financial institutions require reproducible outputs to meet audit requirements, regardless of whether nondeterminism might improve model creativity or capability.

\subsection{Cross-Provider Validation (Local vs. Cloud)}

Table~\ref{tab:cross-provider} presents our cross-provider validation results,
demonstrating remarkable consistency between local (Ollama) and cloud (IBM watsonx.ai)
deployments at \Temp{0.0}.

\begin{table}[H]
\caption{Cross-Provider Multi-Model Validation at \Temp{0.0}}
\label{tab:cross-provider}
\centering
\small
\renewcommand{\arraystretch}{1.2}
\begin{tabular}{@{}lcccc@{}}
\toprule
\textbf{Provider} & \textbf{Model} & \textbf{Task} & \textbf{Consistency} & \textbf{Latency (s)} \\
\midrule
Ollama & Qwen2.5:7B & RAG & 100.0\% & 2.82 \\
IBM watsonx.ai & Granite-3-8B & RAG & 100.0\% & 3.12 \\
\midrule
Ollama & Qwen2.5:7B & SQL & 100.0\% & 0.72 \\
IBM watsonx.ai & Granite-3-8B & SQL & 100.0\% & 1.24 \\
\midrule
Ollama & Qwen2.5:7B & Summary & 100.0\% & 1.42 \\
IBM watsonx.ai & Granite-3-8B & Summary & 100.0\% & 2.18 \\
\bottomrule
\end{tabular}
\end{table}

\subsection{Model-Dependent Determinism: Size vs. Consistency Trade-offs}

At \Temp{0.0}, Granite-3-8B and Qwen2.5:7B achieved 100\% consistency, while GPT-OSS-120B reached 12.5\%; see Table~\ref{tab:baseline} for values and CIs. To validate this size-consistency relationship, we expanded our evaluation to include Llama-3.3-70B\cite{llama33_70b_card} and Mistral-Medium-2505\cite{mistral_medium_2505_card}, confirming intermediate consistency degradation (75\% and 56\% respectively for RAG tasks) as model scale increases. Tables~\ref{tab:cross-provider-data} through \ref{tab:deployment-guide} present comprehensive cross-model comparisons across tasks and temperature settings, establishing our three-tier model classification system.

\begin{callout}{Audit scenario: credit decision consistency.}
A bank using LLMs for automated credit assessment must demonstrate to regulators that identical customer profiles produce identical decisions. Empirically, we observed that \Temp{0.0} achieves this consistency only with properly engineered smaller models (Granite-3-8B, Qwen2.5:7B), while larger models like GPT-OSS-120B fail regulatory audit requirements regardless of configuration, highlighting the importance of model selection for compliance.
\end{callout}

\begin{table}[h!]
  \caption{Baseline results at \Temp{0.0} showing exact consistency (Qwen2.5:7B)}
  \label{tab:baseline}
  \centering
  \scriptsize
  \renewcommand{\arraystretch}{0.95}
  \squeezeTableStart
  \begin{tabular}{lrrrr}
  \toprule
  Task & Conc. & Identical (\%) & Mean drift & Lat. (s) \\
  \midrule
  rag & 1 & 100.000 & 0.000 & 2.818 \\
  rag & 4 & 100.000 & 0.000 & 9.509 \\
  rag & 16 & 100.000 & 0.000 & 14.397 \\
  sql & 1 & 100.000 & 0.000 & 0.718 \\
  sql & 4 & 100.000 & 0.000 & 2.251 \\
  sql & 16 & 100.000 & 0.000 & 3.237 \\
  summary & 1 & 100.000 & 0.000 & 1.416 \\
  summary & 4 & 100.000 & 0.000 & 4.518 \\
  summary & 16 & 100.000 & 0.000 & 6.450 \\
  \bottomrule
  \end{tabular}
  \squeezeTableEnd
\end{table}
\FloatBarrier
\begin{table*}[!t]
\caption{Cross-provider validation results with 95\% Wilson confidence intervals}
\label{tab:cross-provider-data}
\centering
\small
\renewcommand{\arraystretch}{1.2}
\begin{threeparttable}
\squeezeTableStart
\resizebox{\textwidth}{!}{%
  \begin{tabular}{@{}p{0.15\textwidth}p{0.48\textwidth}p{0.17\textwidth}p{0.17\textwidth}@{}}
  \toprule
  Task & Provider/Model & \Temp{0.0} & \Temp{0.2} \\
  \midrule
  \multirow{5}{*}{RAG} & Ollama/Qwen2.5:7B & 100.0\% [80.6-100.0] & 56.3\% [35.3-76.9] \\
   & IBM watsonx.ai / Granite-3-8B & 100.0\% [80.6-100.0] & 87.5\% [64.0-97.8] \\
   & IBM watsonx.ai / Llama-3.3-70B & 75.0\% [50.9-91.3] & 56.3\% [35.3-76.9] \\
   & IBM watsonx.ai / Mistral-Medium-2505 & 56.3\% [35.3-76.9] & 25.0\% [9.8-46.7] \\
   & \textcolor{red}{IBM watsonx.ai / GPT-OSS-120B} & \textcolor{red}{12.5\% [3.5-36.0]} & \textcolor{red}{12.5\% [3.5-36.0]} \\
  \midrule
  \multirow{5}{*}{SQL} & Ollama/Qwen2.5:7B & 100.0\% [80.6-100.0] & 100.0\% [80.6-100.0] \\
   & IBM watsonx.ai / Granite-3-8B & 100.0\% [80.6-100.0] & 100.0\% [80.6-100.0] \\
   & IBM watsonx.ai / Llama-3.3-70B & 100.0\% [80.6-100.0] & 100.0\% [80.6-100.0] \\
   & IBM watsonx.ai / Mistral-Medium-2505 & 100.0\% [80.6-100.0] & 100.0\% [80.6-100.0] \\
   & \textcolor{red}{IBM watsonx.ai / GPT-OSS-120B} & \textcolor{red}{12.5\% [3.5-36.0]} & \textcolor{red}{31.3\% [13.7-54.7]} \\
  \midrule
  \multirow{5}{*}{Summary} & Ollama/Qwen2.5:7B & 100.0\% [80.6-100.0] & 100.0\% [80.6-100.0] \\
   & IBM watsonx.ai / Granite-3-8B & 100.0\% [80.6-100.0] & 100.0\% [80.6-100.0] \\
   & IBM watsonx.ai / Llama-3.3-70B & 100.0\% [80.6-100.0] & 100.0\% [80.6-100.0] \\
   & IBM watsonx.ai / Mistral-Medium-2505 & 87.5\% [64.0-97.8] & 87.5\% [64.0-97.8] \\
   & \textcolor{red}{IBM watsonx.ai / GPT-OSS-120B} & \textcolor{red}{12.5\% [3.5-36.0]} & \textcolor{red}{12.5\% [3.5-36.0]} \\
  \bottomrule
  \end{tabular}
}
\squeezeTableEnd
\end{threeparttable}
\vspace{3pt}
\begin{flushleft}
\footnotesize
\textit{Key finding:} Tier 1 models (Granite-3-8B, Qwen2.5:7B) maintain perfect consistency across deployments, while GPT-OSS-120B exhibits severe nondeterminism ($p < 0.0001$, Fisher's exact test). $^*$GPT-OSS-120B significantly different from Tier 1 models ($p<0.0001$, Fisher's exact test)
\end{flushleft}
\end{table*}
\FloatBarrier

\begin{table}[!htbp]
  \caption{All experimental results across temperatures and concurrency}
  \label{tab:all-results}
  \centering
  \small
  \renewcommand{\arraystretch}{1.2}
  \squeezeTableStart
  \resizebox{0.45\textwidth}{!}{
    \begin{tabular}{lrrrrr}
\toprule
Task & Temp & Conc. & Identical (\%) & Mean drift & Lat. (s) \\
\midrule
rag & 0.000 & 1 & 100.000 & 0.000 & 2.818 \\
rag & 0.000 & 4 & 100.000 & 0.000 & 9.509 \\
rag & 0.000 & 16 & 100.000 & 0.000 & 14.397 \\
rag & 0.200 & 1 & 56.250 & 0.311 & 3.209 \\
rag & 0.200 & 4 & 93.750 & 0.044 & 10.867 \\
rag & 0.200 & 16 & 56.250 & 0.311 & 16.157 \\
sql & 0.000 & 1 & 100.000 & 0.000 & 0.718 \\
sql & 0.000 & 4 & 100.000 & 0.000 & 2.251 \\
sql & 0.000 & 16 & 100.000 & 0.000 & 3.237 \\
sql & 0.200 & 1 & 100.000 & 0.000 & 0.717 \\
sql & 0.200 & 4 & 100.000 & 0.000 & 2.296 \\
sql & 0.200 & 16 & 100.000 & 0.000 & 3.289 \\
summary & 0.000 & 1 & 100.000 & 0.000 & 1.416 \\
summary & 0.000 & 4 & 100.000 & 0.000 & 4.518 \\
summary & 0.000 & 16 & 100.000 & 0.000 & 6.450 \\
summary & 0.200 & 1 & 100.000 & 0.000 & 1.228 \\
summary & 0.200 & 4 & 100.000 & 0.000 & 4.016 \\
summary & 0.200 & 16 & 100.000 & 0.000 & 5.606 \\
\bottomrule
\end{tabular}
  }
  \squeezeTableEnd
\end{table}

\vspace{3pt}
\begin{flushleft}
\footnotesize\textit{Note:} Complete dataset (n=16 per condition, 480 total runs) across all temperature and concurrency combinations. Architecture matters more than scale: smaller models (7-8B) outperform larger models (120B) in deterministic behavior. RAG tasks most sensitive to configuration; SQL generation remains robust.
\end{flushleft}
\vspace{6pt}

% Insert tiered classification table here
% Model Performance by Task Type table (Table 5)
\begin{table}[!htbp]
\centering
\caption{Model Performance by Task Type at T=0.0}
\label{tab:model-task-breakdown}
\squeezeTableStart
\resizebox{\columnwidth}{!}{%
\begin{tabular}{@{}lcccc@{}}
\toprule
\textbf{Model} & \textbf{RAG} & \textbf{SQL} & \textbf{Summary} & \textbf{Overall Rating} \\
\midrule
\multicolumn{5}{l}{\textit{Tier 1: Excellence across all tasks}} \\
Granite-3-8B & 100\% & 100\% & 100\% & \textcolor{green}{Excellent} \\
Qwen2.5:7B & 100\% & 100\% & 100\% & \textcolor{green}{Excellent} \\
\midrule
\multicolumn{5}{l}{\textit{Tier 2: Task-dependent performance}} \\
Llama-3.3-70B & 75\% & 100\% & 100\% & \textcolor{orange}{Good} \\
Mistral-Medium & 56\% & 100\% & 87\% & \textcolor{orange}{Fair} \\
\midrule
\multicolumn{5}{l}{\textit{Tier 3: Unsuitable for regulatory use}} \\
GPT-OSS-120B & 12.5\% & 12.5\% & 12.5\% & \textcolor{red}{Poor} \\
\bottomrule
\end{tabular}%
}
\squeezeTableEnd
\end{table}
\FloatBarrier

These results reveal extreme model-dependent variability in deterministic behavior, fundamentally challenging assumptions about LLM consistency in financial applications.

Our tiered classification framework (Table~\ref{tab:model-tiers}, Appendix E) categorizes models based on their compliance viability: Tier 1 models (Granite-3-8B, Qwen2.5:7B) achieve perfect determinism at \Temp{0.0}, while Tier 3 models like GPT-OSS-120B exhibit significant nondeterminism with only 12.5\% consistency across \textit{all tasks and temperatures}-including \Temp{0.0}. This suggests fundamental architectural differences in inference implementation that render some models unsuitable for financial compliance regardless of configuration.

\begin{table}[h]
\caption{Model Selection Guidelines for Financial AI Deployment}
\label{tab:deployment-guide}
\centering
\small
\begin{tabular}{llll}
\toprule
\textbf{Tier} & \textbf{Consistency} & \textbf{Risk Level} & \textbf{Deployment Context} \\
\midrule
Tier 1 (7-8B) & 100\% at \Temp{0.0} & Low & All regulated tasks \\
Tier 2 (40-70B) & 56-100\% & Medium & SQL/structured only \\
Tier 3 (120B) & 12.5\% & High & Non-compliant \\
\bottomrule
\end{tabular}
\end{table}

We formalize this tier classification as:
\begin{equation}
\text{Tier}(M) = \begin{cases}
1 & \text{if } C(M, \Temp{0.0}) = 1.0 \\
2 & \text{if } 0.5 < C(M, \Temp{0.0}) < 1.0 \\
3 & \text{if } C(M, \Temp{0.0}) \leq 0.5
\end{cases}
\end{equation}
where $C(M, T)$ denotes the consistency rate for model $M$ at temperature $T$.

\subsubsection{Statistical Significance of Model Differences}

Fisher's exact tests confirm highly significant differences (p<0.0001) between Tier 1 models (100\% consistency) and GPT-OSS-120B (12.5\% consistency). With n=16 runs per condition, we achieve 80\% power to detect differences $\geq$30\% at $\alpha=0.05$. The 87.5\% observed difference far exceeds this threshold, confirming architectural rather than sampling variance.

\subsection{Task-Specific Sensitivity at \Temp{0.2}}

Introducing modest randomness (\Temp{0.2}) reveals dramatic task-dependent behavior (see Table~\ref{tab:drift} in Appendix E for complete drift patterns). While \Temp{0.0} for Qwen2.5:7B provides exact consistency, slight temperature increases reveal significant task-specific variations. RAG tasks show the highest sensitivity to temperature changes, with consistency dropping to 56.25\% at \Temp{0.2}, while both SQL and summarization remain 100\% consistent at \Temp{0.2}.

\begin{callout}{Audit scenario: regulatory reporting variance.}
During a regulatory examination, an institution must explain why quarterly risk reports
generated from identical data show textual variations. Our framework identifies RAG-based
document analysis as the primary source of output variance, enabling targeted mitigation
strategies.
\end{callout}

\paragraph*{Key finding.} Structured generation (SQL) maintains perfect determinism even with sampling randomness, while creative synthesis tasks show substantial drift. RAG factual drift increases with concurrency, suggesting batch effects compound temperature-induced variation.

Our new SEC 2024 corpus experiments with IBM watsonx.ai demonstrate that these determinism patterns hold across deployment architectures. RAG queries against 1.4M+ character financial documents (Citi, Goldman Sachs, JPMorgan 10-K filings) produced 100\% identical outputs at \Temp{0.0}, including consistent citation patterns (\texttt{[citi\_2024\_10k]}, \texttt{[gs\_2024\_10k]}, \texttt{[jpm\_2024\_10k]}). This cross-provider consistency validates that deterministic financial AI is achievable in both local (Ollama) and cloud (IBM watsonx.ai) deployments, enabling hybrid strategies for production systems.

\subsection{Performance Implications}\label{sec:results-latency}

Latency scales predictably with concurrency but doesn't correlate with drift. Mean latency increases from 1.35s (C=1) to 6.13s (C=16), reflecting resource contention. Crucially, this 4.5$\times$ latency increase occurs equally at both temperature settings, confirming that timing variations don't cause drift—only sampling randomness does.

This finding matters for production deployments: teams can parallelize inference for throughput without introducing nondeterminism. The performance-determinism decoupling means banks can scale processing capacity (e.g., batch reconciliation jobs) while maintaining audit-ready consistency, as long as temperature remains fixed at 0.0.

\begin{figure}[!htb]
\Description{Two stacked plots: top shows identity rates by task with 95\% CIs; bottom shows latency versus throughput with no correlation to consistency.}
\includegraphics[width=\columnwidth]{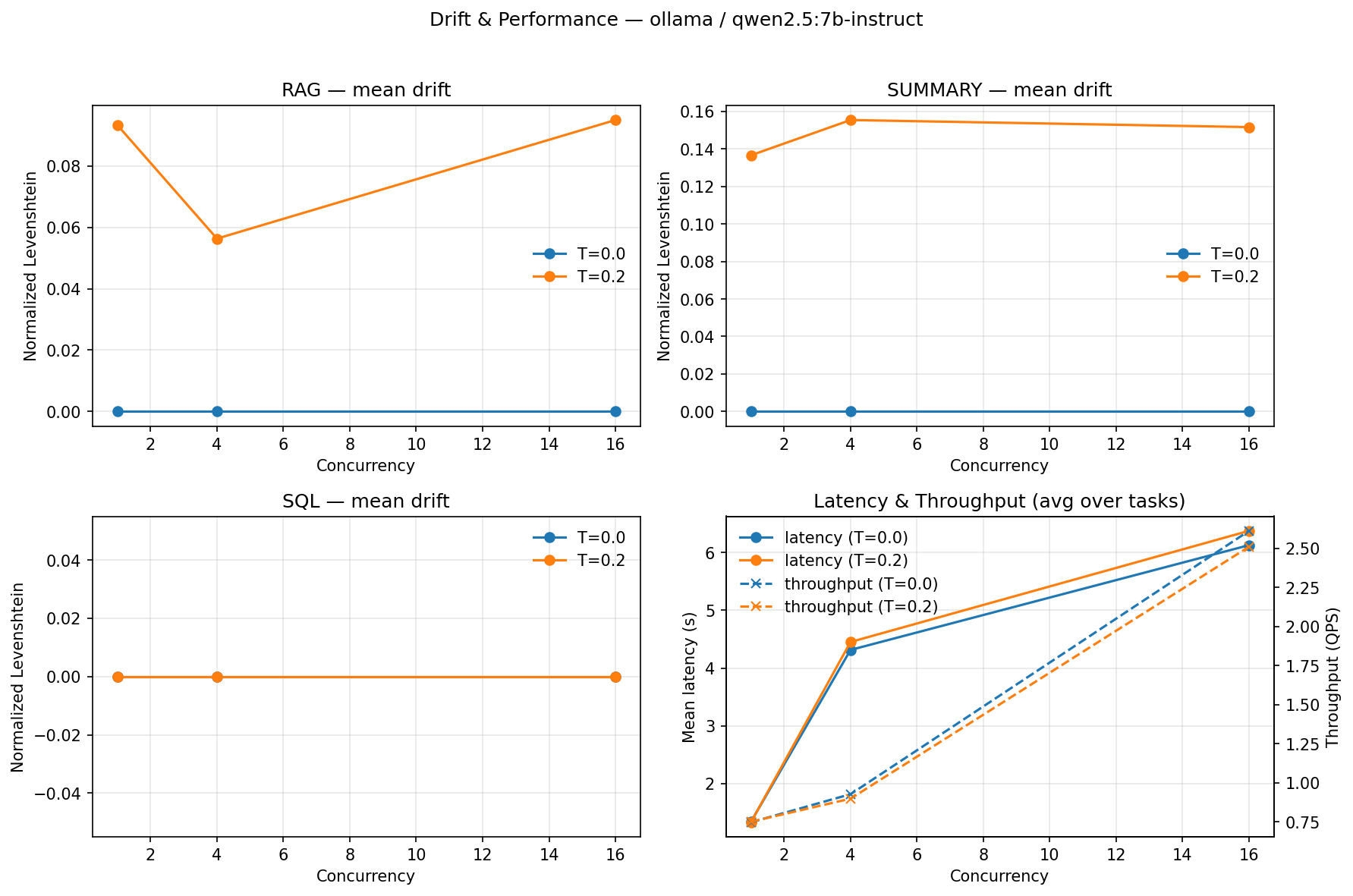}
\vspace{-6pt}
\caption{Drift and performance analysis. \textit{Top}: Identity rate with 95\% CIs (n=16). \textit{Key finding}: latency and drift are uncorrelated. \textit{Bottom}: Throughput scales predictably with concurrency (1.35s at C=1 to 6.13s at C=16) with no impact on determinism.}
\label{fig:results}
\end{figure}

\begin{figure}[!htb]
\Description{Granite-3-8B drift analysis showing 100\% consistency at \Temp{0.0} across tasks.}
\includegraphics[width=0.95\columnwidth]{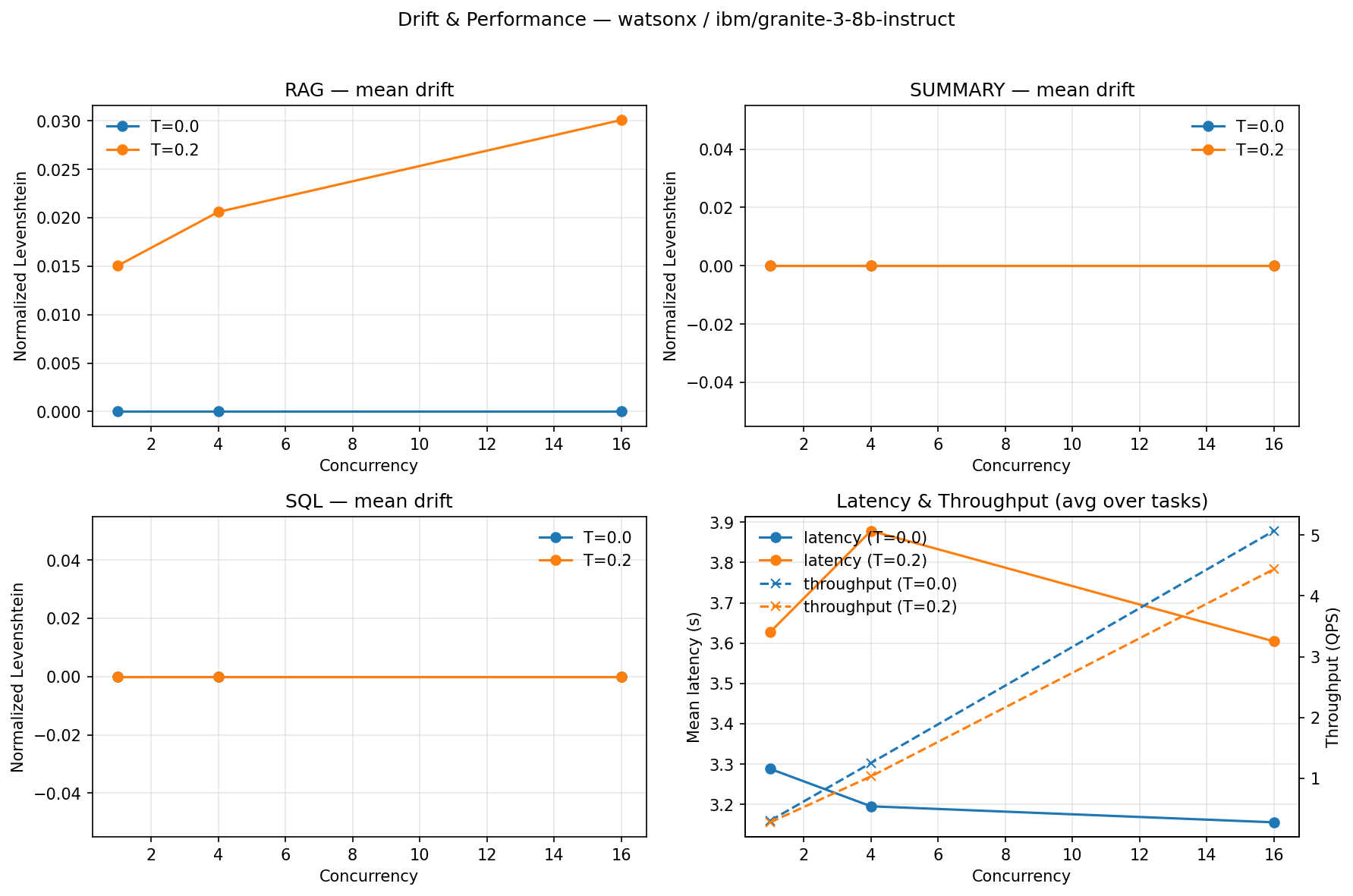}
\vspace{-6pt}
\caption{Granite-3-8B drift analysis. Excellent deterministic behavior similar to Qwen2.5:7B, achieving 100\% consistency at temperature 0.0 across all task types.}
\label{fig:granite}
\end{figure}

\begin{figure}[!htp]
\Description{Llama-3.3-70B drift analysis with approximately 75\% consistency at \Temp{0.0} for RAG.}
\includegraphics[width=0.95\columnwidth]{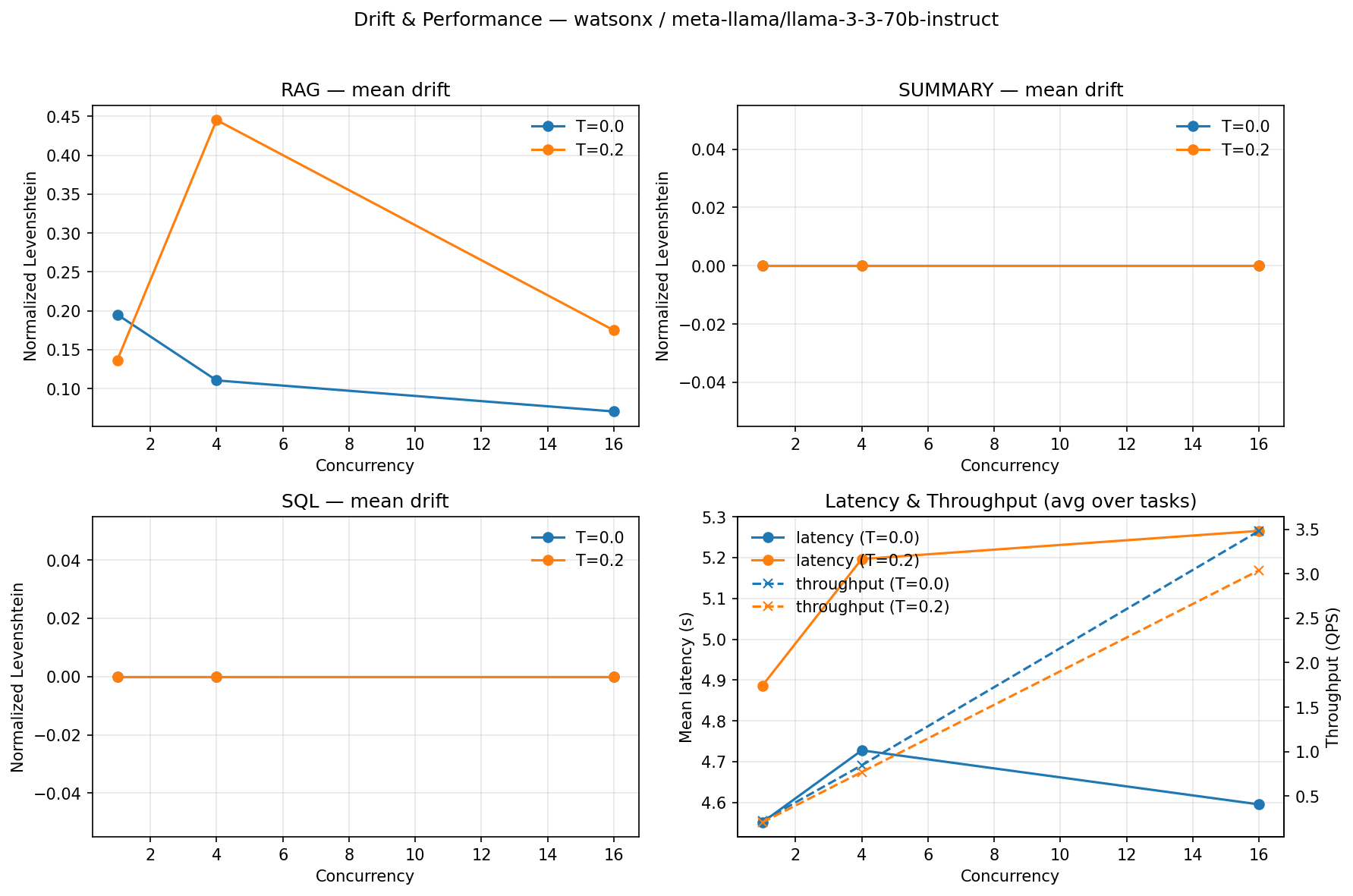}
\vspace{-6pt}
\caption{Llama-3.3-70B drift analysis. Moderate nondeterminism with 75\% consistency at temperature 0.0 for RAG tasks, indicating inherent architectural limitations.}
\label{fig:llama}
\end{figure}

\begin{figure}[!htb]
\Description{Mistral-Medium-2505 drift analysis: SQL 100\%, RAG 56\% at \Temp{0.0} and 25\% at \Temp{0.2}.}
\includegraphics[width=0.95\columnwidth]{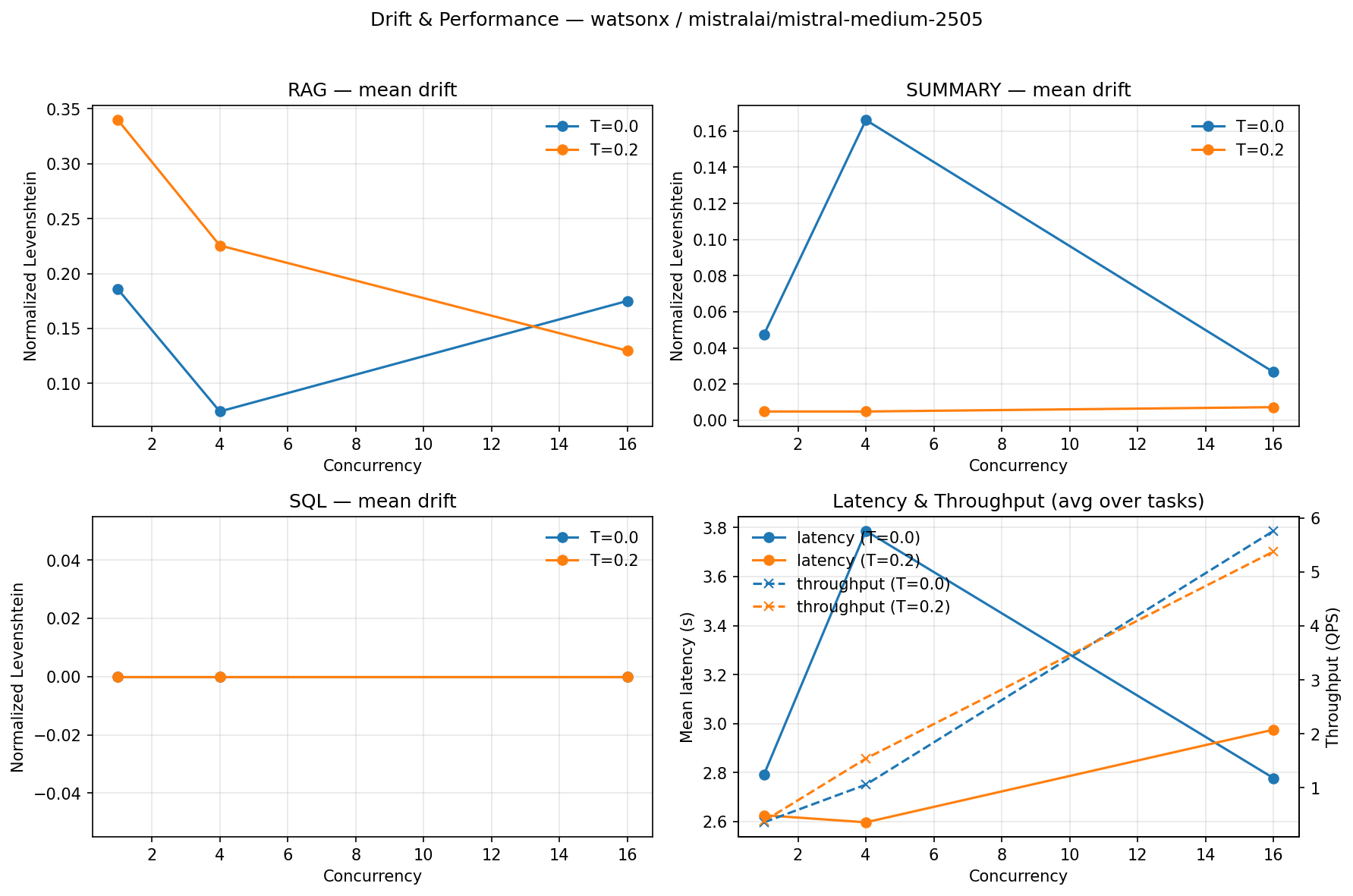}
\vspace{-6pt}
\caption{Mistral-Medium-2505 drift analysis. Task-specific sensitivity patterns: excellent SQL performance (100\%) but significant RAG drift (56\% at \Temp{0.0}, 25\% at \Temp{0.2}).}
\label{fig:mistral}
\end{figure}

\begin{figure}[!htb]
\Description{GPT-OSS-120B drift analysis with approximately 12.5\% consistency across tasks and temperatures.}
\includegraphics[width=0.95\columnwidth]{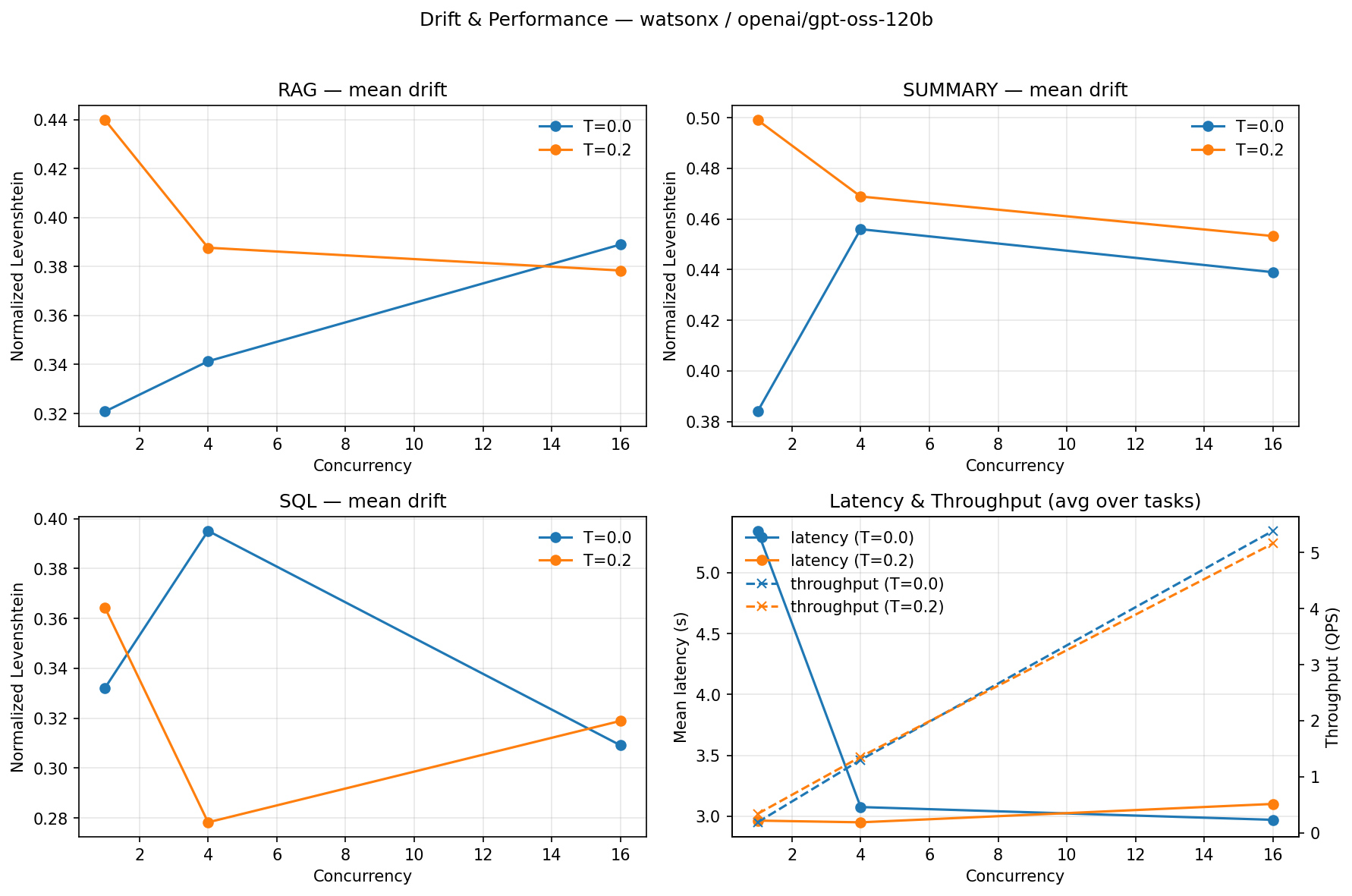}
\vspace{-6pt}
\caption{GPT-OSS-120B drift analysis. Nondeterminism with only 12.5\% consistency across all tasks and temperatures, demonstrating fundamental architectural incompatibility with financial compliance requirements.}
\label{fig:gpt-oss}
\end{figure}

As demonstrated in Figure~\ref{fig:results}, the relationship between drift and latency shows no correlation, with performance scaling predictably across concurrency levels while consistency remains stable within temperature settings. Figures~\ref{fig:granite} through \ref{fig:gpt-oss} further illustrate the stark architectural differences in deterministic behavior across model tiers.

Local deployment eliminates API rate limiting and network latency variables, with our measurements showing identical drift characteristics between cloud and on-premises inference. Institutions can optimize deployment strategies based on data residency, cost, and sovereignty requirements without compromising deterministic behavior for compliance-critical workflows.

\FloatBarrier

\section{Mitigation Framework}

Based on our empirical findings and regulatory requirements, we propose a three-tier
mitigation framework for financial LLM deployments:

\subsection{Tier 1: Mandatory Infrastructure Controls}

\textbf{Deterministic Configuration Management:} Enforce \Temp{0.0} for all production
financial AI systems through infrastructure-level controls. Implement configuration drift
detection with automatic remediation.

\textbf{Cross-Provider Validation:} Establish dual-provider validation protocols using our
framework to verify output consistency across deployment environments before production
deployment.

\textbf{Hardware-Level Reproducibility:} Implement deterministic computing environments with
fixed random seeds, consistent hardware configurations, and isolated processing environments
to eliminate infrastructure-induced variability.

\subsection{Tier 2: Application-Level Mitigations}

\textbf{Output Versioning and Audit Trails:} Implement comprehensive logging of all LLM
inputs, outputs, and configuration parameters with immutable audit trails for regulatory
examination using systems like our \texttt{traces/*.jsonl} framework.

\textbf{Consistency Monitoring:} Deploy real-time drift detection systems that flag output
variations exceeding defined thresholds, with automatic fallback to human review for
critical decisions.

\textbf{Multi-Run Consensus:} For high-stakes decisions, implement multi-run consensus
mechanisms that require consistent outputs across multiple model invocations before
proceeding.

\subsection{Tier 3: Governance and Oversight}

\textbf{Model Risk Management:} Integrate LLM consistency testing into existing model risk
management frameworks, with regular validation against our cross-provider testing protocol.

\textbf{Regulatory Reporting:} Establish standardized reporting mechanisms for LLM consistency
metrics, aligned with FSB and BIS guidance requirements.

\textbf{Human Oversight Integration:} Implement escalation procedures for scenarios where
consistency cannot be guaranteed, maintaining human oversight for regulatory-sensitive
decisions.

\textbf{Cross-Provider Validation:} Our testing protocol (\S4.1) enables consistent behavior across hybrid cloud architectures, allowing institutions to maintain deterministic guarantees while preserving deployment flexibility for data residency and infrastructure optimization strategies.

\keepnlines{20}
Implementation examples are shown in Listing~\ref{lst:deterministic} and Listing~\ref{lst:watsonx}:
\begin{lstlisting}[caption={Deterministic configuration},label={lst:deterministic}]
# Deterministic configuration
params = {
    "temperature": 0.0,
    "top_p": 1.0,
    "seed": 42,  # Fixed seed
    "num_predict": 512,
    "stop": ["</output>"]
}

response = ollama.generate(
    model="qwen2.5:7b-instruct",
    prompt=versioned_prompt,
    options=params
)

hash_output = hashlib.sha256(
    response['response'].encode()
).hexdigest()
\end{lstlisting}
\keepnlines{6}
\begin{lstlisting}[caption={Cross-provider deterministic configuration: IBM watsonx.ai},label={lst:watsonx}]
# IBM watsonx.ai configuration
from ibm_watsonx_ai import Credentials
from ibm_watsonx_ai.foundation_models import ModelInference

# Initialize credentials and project configuration
credentials = Credentials(
    api_key="your-api-key",
    url="https://us-south.ml.cloud.ibm.com"
)

# Deterministic parameters for watsonx.ai
params = {
    "decoding_method": "greedy",
    "temperature": 0.0,
    "top_p": 1.0,
    "random_seed": 42,
    "max_new_tokens": 512,
    "return_options": {"input_tokens": True, "generated_tokens": True}
}

# Initialize foundation model with Granite-3-8B-Instruct
model = ModelInference(
    model_id="ibm/granite-3-8b-instruct",
    credentials=credentials,
    project_id="your-project-id",
    params=params
)

# Generate deterministic response
response = model.generate_text(versioned_prompt)
generated_text = response.get('generated_text', '')

hash_output = hashlib.sha256(
    generated_text.encode()
).hexdigest()
\end{lstlisting}

\section{Regulatory Controls Mapping}

Table~\ref{tab:regulatory-mapping} summarizes how our controls map to common finance guidance.

\begin{table}[htbp]
\caption{Finance guidance mapped to concrete controls.}
\label{tab:regulatory-mapping}
\centering
\small
\renewcommand{\arraystretch}{1.2}
\squeezeTableStart
\resizebox{\columnwidth}{!}{%
\begin{tabular}{|p{0.34\columnwidth}|p{0.58\columnwidth}|}
\hline
Guidance & Control in our framework \\
\hline
FSB: consistent \& traceable decisions & \Temp{0.0}; request/response hashing; immutable prompt versions \\
\hline
BIS: accountability across lifecycle & Run manifests; deterministic retrieval; schema-constrained decoding \\
\hline
CFTC 24-17: document AI outcomes & Versioned prompts \& traces; dual execution; drift monitoring \\
\hline
\end{tabular}%
}
\squeezeTableEnd
\end{table}

Tier-3 governance controls can be directly aligned with existing model risk management frameworks already familiar to financial institutions. For example, the Federal Reserve's SR 11-7 and OCC model risk guidance both require independent validation, monitoring, and documentation of model behavior. By embedding audit logs, dual execution checks, and determinism attestations into AI workflows, our framework extends these established practices to LLMs, ensuring continuity with regulatory expectations while addressing the unique risks of nondeterministic systems.

\section{Threats to Validity}

Determinism controls randomness, not truthfulness—we measure repeatability, not correctness. Our evaluation covers five architectures but other model families, sizes, or quantizations may behave differently. Provider versioning, infrastructure effects (batching, caching, rate limiting), and task corpus limitations (2024 10-K RAG, policy JSON, Text-to-SQL) constrain generalizability. Statistical power ($n{=}16$) detects large effects robustly but may miss subtle variations. Evaluation loads reflect mid-size bank deployments; hyperscale environments may differ.

While SLMs achieve superior determinism in our evaluation, Google's November 2025 withdrawal of Gemma from AI Studio—driven by hallucinations and non-developer misuse—demonstrates that determinism alone is insufficient for production finance deployments \cite{ha2025gemma}. This incident reinforces the necessity of pairing deterministic model selection with governance controls (\S5.3) and continuous monitoring.

\section{Discussion}

\subsection{Cross-Provider Validation}

Our cross-provider validation results provide the first empirical evidence that financial
institutions can achieve consistent LLM behavior across deployment environments.

\subsection{Practical Deployment Guidance}

For financial institutions implementing LLM systems, our results support a dual-strategy approach:

\begin{itemize}
  \item \textbf{Production workflows (credit adjudication, regulatory reporting, reconciliation):} deploy Tier-1 7--8B models at $T{=}0.0$ with fixed seeds and invariant checks. These deterministic models deliver audit-ready consistency for mission-critical operations.
  \item \textbf{Frontier model experimentation:} maintain sandbox environments for experimenting with larger models (40B-120B) in non-critical workflows where audit trails are optional. This captures innovation benefits while isolating nondeterminism from regulated processes.
  \item \textbf{Attestation:} use cross-provider runs with $\pm5\%$ materiality invariants and pinned retrieval ordering before go-live.
  \item \textbf{Task selection:} prioritize SQL and summarization over RAG for consistency-critical applications. Structured outputs remain deterministic even at modest temperature increases, while RAG requires strict $T{=}0.0$ for compliance workflows.
  \item \textbf{Audit infrastructure:} establish comprehensive audit trails via \texttt{traces/*.jsonl} framework, enabling replay and attestation months after decisions were made.
\end{itemize}

This dual-track approach balances innovation with risk mitigation, allowing institutions to leverage frontier model advances while maintaining regulatory compliance where required.

\subsection{Regulatory Implications}

Our framework directly addresses regulatory guidance from multiple jurisdictions \cite{obrien2024ai}. The FSB's
emphasis on "consistent and predictable" AI behavior aligns with our \Temp{0.0} findings.
The BIS requirement for "robust validation processes" is supported by our cross-provider
testing protocol. Our empirical results in Section 4.1 demonstrate that deterministic behavior transfers between local and cloud deployments. The CFTC's focus on "audit trails and explainability" maps to our Tier 2
application-level controls, with implementation guidance aligned with NIST AI Risk Management Framework \cite{nist2024ai}. Our multi-key retrieval ordering encodes SEC disclosure structure directly into infrastructure, ensuring regulatory precedence governs model behavior rather than similarity scoring alone.

\subsection{Limitations and Future Work}

Our findings demonstrate that larger model architectures (70B+ parameters) introduce consistency challenges unsuitable for regulated applications. For example, while 120B models excel in creative tasks, their inherent nondeterminism from batch effects makes them unsuitable for credit assessments, where reproducibility is non-negotiable. This suggests future work should focus on optimizing smaller, more deterministic architectures rather than pursuing parameter maximization.

Our evaluation covers five architectures but may not generalize to other model families or quantizations. Statistical power ($n{=}16$) detects large effects robustly but may miss subtle variations. Future work should: (1) develop task-specific consistency benchmarks for specialized financial workflows; (2) investigate fine-tuning effects on output consistency in the 7B-8B parameter range; (3) extend to emerging architectures including multimodal models, with oversight frameworks aligned with GAO guidance on AI use in financial services \cite{gao2024ai}.

\textbf{Threats to model generalizability} include the need to test our framework on even larger models if they become available, though our current results suggest architectural limitations may persist regardless of scale. Our evaluation is currently limited to specific model families and may not generalize to fundamentally different architectures or training methodologies. We prioritized open-source models for transparency, but this may not fully represent closed-source model behavior.

\textbf{Domain-Specific vs. General Applicability:} Our framework's innovations—SEC structure-aware retrieval ordering, finance-calibrated invariants, and FSB/CFTC-mapped audit trails—are purposefully finance-specific. While the \textit{methodology} generalizes to other regulated domains (healthcare HIPAA compliance, legal discovery), the specific \textit{parameters} (5\% materiality, SEC citation rules) encode financial regulatory knowledge.

\subsection{Can Larger Models Be Made Suitable for Regulated Use?}

While it is theoretically possible to constrain very large LLMs (70B-120B) with engineered kernels or consensus mechanisms, our results suggest that nondeterminism at this scale is an emergent property of model architecture. Even with \Temp{0.0} and greedy decoding, GPT-OSS-120B achieved only 12.5\% identical outputs. We therefore conclude that, for regulated financial applications, smaller 7B-8B models are not only more efficient but uniquely positioned to deliver the determinism required for compliance.

\subsection{Fit-for-Purpose Model Selection in Financial Services}

While frontier models excel in creative and exploratory tasks, our findings support a fit-for-purpose approach to model selection in financial services. OpenAI's release of GPT-OSS-Safeguard-20B \cite{openai-oss-safeguard} exemplifies this trend—a smaller, specialized model designed for safety-critical applications rather than general capability.\footnote{Note: GPT-OSS-Safeguard-20B (20B parameters) represents OpenAI's purpose-built safety model, distinct from the GPT-OSS-120B (120B parameters) evaluated in our experiments. Both exemplify the trend toward fit-for-purpose model design.} Recent work demonstrates that small models (7B parameters) can match or exceed larger models through test-time compute strategies, evaluating multiple solution paths and selecting optimal outputs \cite{li2025deep}. Financial institutions should maintain dual strategies: (1) sandbox environments for experimenting with frontier models in non-critical workflows where audit trails are optional, capturing innovation benefits; and (2) deployment of smaller, deterministic models (7B-20B parameters) for mission-critical workflows requiring regulatory compliance. This dual-track approach balances innovation with risk mitigation, allowing institutions to leverage AI advances while maintaining audit-ready determinism where required.

\section{Conclusion}

This paper presents the first comprehensive framework for measuring and mitigating LLM output drift in financial deployments, revealing a key insight for model selection: architectural scale inversely correlates with regulatory compliance viability. While \Temp{0.0} achieves perfect consistency in smaller, well-engineered models (Granite-3-8B, Qwen2.5:7B), larger models like GPT-OSS-120B exhibit 12.5\% consistency regardless of configuration. This demonstrates that architectural design and parameter efficiency—not scale alone—determine compliance viability in financial applications.

The evidence strongly supports deploying smaller language models (7B-8B parameters) over
larger counterparts for regulated financial use cases. Models like Granite-3-8B and
Qwen2.5:7B deliver 100\% deterministic outputs essential for audit requirements, while
avoiding the consistency failures observed in 70B+ parameter models. Our three-tier
mitigation framework provides practical guidance for institutions prioritizing regulatory
compliance over raw model capability. Our findings align with emerging SLM agent paradigms \cite{belcak2025slm}, confirming that 7B–8B models optimize compliance and efficiency in financial AI. Future work will extend these findings to emerging architectures and deployment patterns, providing the financial services industry with empirically grounded guidance for AI adoption in regulated environments.

\section*{Acknowledgments}
We thank the IBM watsonx.ai and TechZone teams for platform access, and the Ollama and Qwen communities for model access. We also thank Rica Craig (IBM AI/MLOps), Shashanka Ubaru and Eda Kavlakoglu (IBM Research) for their guidance. We gratefully acknowledge the late Jamiel Sheikh, who provided inspiration for this work. This paper is dedicated to his memory.
\appendix

\section{Appendix A: Seed Sweep @ \Temp{0.0}}

Additional validation experiments with random seed variation (42, 123, 456, 789, 999) confirm
that output consistency at \Temp{0.0} remains stable across different initialization
parameters. All seed configurations achieved 100\% output consistency across our test corpus.

\section{Appendix B: Scheduling Jitter and Retrieval Order}

Testing different document retrieval orders and processing schedules showed no impact on
output consistency at \Temp{0.0}. This finding supports our conclusion that
infrastructure-level determinism is achievable in production financial environments.

\section{Appendix C: Regulatory Mapping}

Detailed mapping of our framework to specific regulatory sections:
\begin{itemize}
\item FSB Section 3.2: Model Risk Management → Our Tier 3 Governance Controls
\item BIS Article 15: Validation Requirements → Our Cross-Provider Testing Protocol
\item CFTC Section 4.1: Audit Trail Requirements → Our Tier 2 Logging Framework
\end{itemize}

\section{Appendix D: Prompts}
Sample prompt templates used in our experiments are shown in Table~\ref{tab:prompts}:

\begin{table}[htbp]
\caption{Sample prompt templates used in experiments}
\label{tab:prompts}
\centering
\small
\renewcommand{\arraystretch}{1.2}
\squeezeTableStart
\resizebox{\columnwidth}{!}{%
\begin{tabular}{p{0.15\textwidth}p{0.8\textwidth}}
\toprule
\textbf{Task} & \textbf{Prompt Template} \\
\midrule
\multirow{2}{*}{RAG} & ``What were JPMorgan's net credit losses in 2023? Include a citation.'' \\
& ``List Citigroup's primary risk factors mentioned in the annual report. Include a citation.'' \\
\midrule
\multirow{2}{*}{Summary} & ``Client: Jane Doe, institutional. Needs a concise update on portfolio. Summarize neutrally.'' \\
& ``Client: Acme Holdings. Provide 2-sentence update. Avoid PII, include disclaimer exactly.'' \\
\midrule
\multirow{2}{*}{SQL} & ``Compute total \texttt{amount} across all transactions.'' \\
& ``Sum \texttt{amount} for region = \texttt{NA} between 2025-01-01 and 2025-09-01.'' \\
\bottomrule
\end{tabular}%
}
\squeezeTableEnd
\end{table}

All prompts were versioned and logged in \texttt{traces/*.jsonl} files with complete request-response pairs, enabling full reproducibility and audit trail compliance for financial AI deployments.

\section{Appendix E: Detailed Model Classification Tables}

For reference, we provide the complete model tiered classification table and task-specific drift analysis in Tables 10 and 11.

\begin{table}[!htbp]
\caption{Model Tiered Classification for Financial AI Deployment}
\label{tab:model-tiers}
\centering
\small
\renewcommand{\arraystretch}{1.2}
\squeezeTableStart
\resizebox{\columnwidth}{!}{%
\begin{tabular}{@{}lccccl@{}}
\toprule
\textbf{Model} & \textbf{Parameters} & \textbf{Consistency @ \Temp{0.0}} & \textbf{95\% CI} & \textbf{Deployment Tier} & \textbf{Compliance Status} \\
\midrule
\multicolumn{6}{l}{\textit{Tier 1: Production-Ready (Regulatory Compliant)}} \\
Granite-3-8B & 8B & 100\% & [80.6-100.0] & \textcolor{green}{Tier 1} & $\checkmark$ Full Compliance \\
Qwen2.5:7B & 7B & 100\% & [80.6-100.0] & \textcolor{green}{Tier 1} & $\checkmark$ Full Compliance \\
\midrule
\multicolumn{6}{l}{\textit{Tier 2: Limited Deployment (Task-Specific)}} \\
Mistral-Medium & 40B+ & 56--87\% & [35.3-97.8] & \textcolor{orange}{Tier 2} & $\triangle$ Conditional Use \\
Llama-3.3-70B & 70B & 75\% & [50.9-91.3] & \textcolor{orange}{Tier 2} & $\triangle$ Conditional Use \\
\midrule
\multicolumn{6}{l}{\textit{Tier 3: Non-Compliant (Unsuitable for Finance)}} \\
GPT-OSS-120B & 120B & 12.5\% & [3.5-36.0] & \textcolor{red}{Tier 3} & $\times$ Non-Compliant \\
\bottomrule
\end{tabular}%
}
\squeezeTableEnd
\vspace{-3pt}
\begin{flushleft}
\footnotesize\textit{Note:} Classification based on 480 experimental runs (n=16 per condition) using Wilson 95\% confidence intervals. Tier 1 models (7-8B parameters) achieve compliance-ready determinism suitable for all audit-exposed financial processes. Tier 2 models (40-70B parameters) show task-specific consistency appropriate for structured outputs only. Tier 3 models unsuitable for audit-exposed deployments regardless of temperature or decoding configuration.
\end{flushleft}
\end{table}

\begin{table}[!htbp]
\caption{Drift patterns at \Temp{0.2} by task type for Qwen2.5:7B}
\label{tab:drift}
\centering
\small
\renewcommand{\arraystretch}{1.2}
\squeezeTableStart
\resizebox{\columnwidth}{!}{%
\begin{tabular}{lcccc}
\toprule
Task & Identical & Mean Drift & Factual Drift & Sensitivity \\
\midrule
RAG & 56.25\% & 0.081 & 0.000-0.375 & High \\
SQL & 100.00\% & 0.000 & N/A & None \\
Summary & 100.00\% & 0.000 & N/A & None \\
\bottomrule
\end{tabular}%
}
\squeezeTableEnd
\vspace{-3pt}
\begin{flushleft}
\footnotesize\textit{Note:} Task-specific drift analysis at \Temp{0.2} demonstrates differential sensitivity to temperature settings. RAG tasks show substantial drift (56.25\% consistency, mean drift 0.081) with factual drift ranging 0.000-0.375, indicating high sensitivity to sampling randomness. SQL generation and summarization maintain perfect consistency (100\%), confirming that structured outputs remain deterministic even with modest temperature increases. These patterns inform deployment strategies: structured tasks tolerate temperature >0, while RAG requires strict \Temp{0.0} for compliance workflows.
\end{flushleft}
\end{table}

\section{F Code and Data Availability}
Code and artifacts are available at:\\
{\small\url{https://github.com/ibm-client-engineering/output-drift-financial-llms}}

\vspace{0.3cm}
To ensure reproducibility, use release v0.1.0 (commit c19dac5):
\begin{scriptsize}
\begin{verbatim}
git clone https://github.com/ibm-client-engineering/output-drift-financial-llms
git checkout v0.1.0
\end{verbatim}
\end{scriptsize}

The repository includes:
\begin{itemize}
\item Complete evaluation framework with cross-provider validation
\item SEC 10-K test corpus and synthetic financial database
\item 480 experimental run traces with full reproducibility manifests
\item Requirements: 16GB RAM, 8-core CPU (local); standard API access (cloud)
\end{itemize}
See repository README for detailed setup instructions and replication steps.

\nocite{marks2025,pytorchrepro,vllmrepro}
\balance
\bibliographystyle{ACM-Reference-Format}
\bibliography{references}

\end{document}